\documentclass[sigconf]{acmart}

\usepackage[linesnumbered,commentsnumbered,ruled,vlined]{algorithm2e}
\usepackage{multirow}
\usepackage{multicol}
\usepackage{graphicx}
\usepackage{subfigure}
\newcommand{\tabincell}[2]{\begin{tabular}{@{}#1@{}}#2\end{tabular}}

\AtBeginDocument{%
  \providecommand\BibTeX{{%
    \normalfont B\kern-0.5em{\scshape i\kern-0.25em b}\kern-0.8em\TeX}}}

\begin{document}
\copyrightyear{2021}
\acmYear{2021}
\setcopyright{acmcopyright}\acmConference[SIGSPATIAL '21]{29th International Conference on Advances in Geographic Information Systems}{November 2--5, 2021}{Beijing, China}
\acmBooktitle{29th International Conference on Advances in Geographic Information Systems (SIGSPATIAL '21), November 2--5, 2021, Beijing, China}
\acmPrice{15.00}
\acmDOI{10.1145/3474717.3484212}
\acmISBN{978-1-4503-8664-7/21/11}

%%
%% The "title" command has an optional parameter,
%% allowing the author to define a "short title" to be used in page headers.

\title{Automated Feature-Topic Pairing:  Aligning Semantic and Embedding Spaces in Spatial Representation Learning}

\author{Dongjie Wang}
\email{wangdongjie@Knights.ucf.edu}
\affiliation{%
  \institution{University of Central Florida}
  \streetaddress{}
  \city{Orlando}
  \state{Florida}
  \country{United States}
  \postcode{}
}

\author{Kunpeng Liu}
\email{kunpengliu@knights.ucf.edu}
\affiliation{%
  \institution{University of Central Florida}
  \streetaddress{}
  \city{Orlando}
  \state{Florida}
  \country{United States}
  \postcode{}
}

\author{David Mohaisen}
\email{mohaisen@ucf.edu}
\affiliation{%
  \institution{University of Central Florida}
  \streetaddress{}
  \city{Orlando}
   \state{Florida}
  \country{United States}
  \postcode{}
}

\author{Pengyang Wang}
\email{pywang@um.edu.mo}
\affiliation{%
  \institution{University of Macau}
  \streetaddress{}
  \city{Macau}
  \state{}
  \country{China}
  \postcode{}
}

\author{Chang-Tien Lu}
\email{ctlu@vt.edu}
\affiliation{%
  \institution{Virginia Tech}
  \streetaddress{}
  \city{}
  \state{Virginia}
  \country{United States}
  \postcode{}
}

\author{Yanjie Fu}
\authornote{Concat Author.}
\email{yanjie.fu@ucf.edu}
\affiliation{%
  \institution{University of Central Florida}
  \streetaddress{}
  \city{Orlando}
  \state{Florida}
  \country{United States}
  \postcode{}
}

\ccsdesc[500]{Computing methodologies~Learning latent representations}

\keywords{spatial representation learning, multiple spaces alignment}

% \ccsdesc[500]{Computer systems organization~Embedded systems}
% \ccsdesc[300]{Computer systems organization~Redundancy}
% \ccsdesc{Computer systems organization~Robotics}
% \ccsdesc[100]{Networks~Network reliability}

%%
%% Keywords. The author(s) should pick words that accurately describe
%% the work being presented. Separate the keywords with commas.

%% A "teaser" image appears between the author and affiliation
%% information and the body of the document, and typically spans the
%% page.
% \begin{teaserfigure}
%   \includegraphics[width=\textwidth]{sampleteaser}
%   \caption{Seattle Mariners at Spring Training, 2010.}
%   \Description{Enjoying the baseball game from the third-base
%   seats. Ichiro Suzuki preparing to bat.}
%   \label{fig:teaser}
% \end{teaserfigure}

%%
%% This command processes the author and affiliation and title
%% information and builds the first part of the formatted document.
\renewcommand{\shortauthors}{Wang, et al.}

\begin{abstract}

Automated characterization of spatial data is a kind of critical  geographical intelligence. 
As an emerging technique for characterization, Spatial Representation Learning (SRL) uses deep neural networks (DNNs) to learn non-linear embedded features of spatial data for characterization. However, SRL extracts features by internal layers of DNNs, and thus suffers from lacking semantic labels. Texts of spatial entities, on the other hand, provide semantic understanding of latent feature labels, but is insensible to deep SRL models. 
How can we teach a SRL model to discover appropriate topic labels in texts and pair learned features with the labels?
This paper formulates a new problem: feature-topic pairing, and proposes a novel Particle Swarm Optimization (PSO) based deep learning framework.
Specifically, we formulate the feature-topic pairing problem into an automated alignment task between 1) a latent embedding feature space and 2) a textual semantic topic space. 
We decompose the alignment of the two spaces into : 1) point-wise alignment, denoting the correlation between a topic distribution and an embedding vector; 2) pair-wise alignment, denoting the consistency between a feature-feature similarity matrix  and a topic-topic similarity matrix.
We  design a PSO based solver to simultaneously select an optimal set of topics and learn corresponding features based on the selected topics. 
We develop a closed loop algorithm to iterate between 1) minimizing losses of representation reconstruction and feature-topic alignment and  2) searching the best topics.  
%This framework can be generalized to not just spatial data, but also other feature learning problems with generic graphs and texts. 
Finally, we present extensive experiments to demonstrate the enhanced performance of our method. 
\end{abstract}

\maketitle

\section{Introduction}
Spatial representation learning (SRL) refers to exploiting representation learning techniques to learn features of spatial network data, which has been successfully applied in many real-world scenarios, such as transportation networks, power networks,  social networks, water supply networks~\cite{zhang2018network}. 
In reality, many practical applications  need to understand not just which features are effective, but also what these  effective features stand for.  
This issue relates to two tasks: 1) deep representation learning; 2) label generation and matching for latent embedded features.
Although there has been a rich body of work in SRL, including node embedding, autoencoder, random walk, adversarial learning, generative learning based methods with spatial data~\cite{wang2017region,wang2018you, wang2018learning,wang2020defending}, research in unifying the two tasks is still in its early stage.

\begin{figure}[!t]
	\centering
	\includegraphics[width=0.9\linewidth]{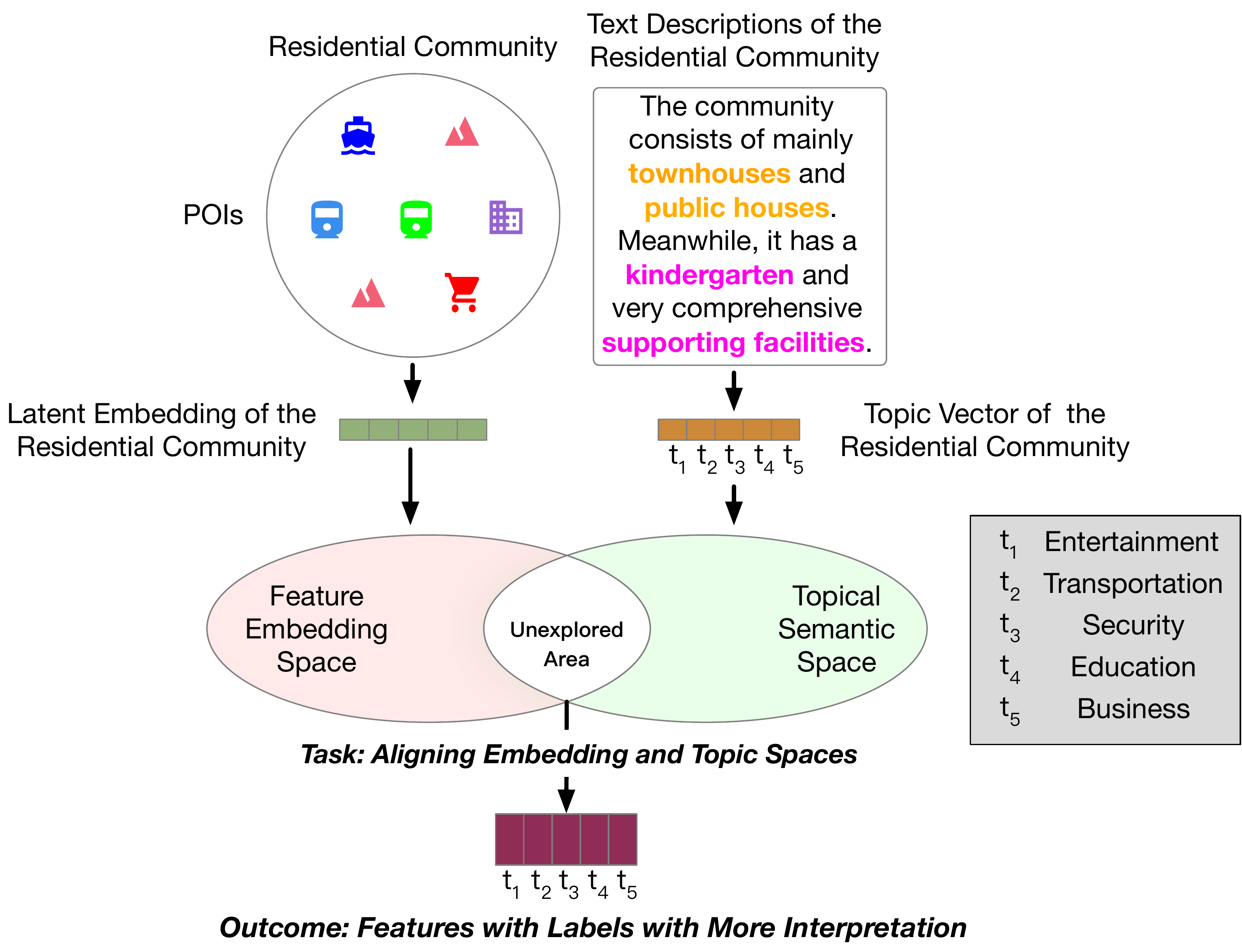}
	\vspace{-0.3cm}
	\captionsetup{justification=centering}
	\caption{The motivation of the feature-topic pairing problem: bridging the gap between feature embedding space and topic semantic space in representation learning. }
	\vspace{-0.6cm}
	\label{fig:motivation}
\end{figure}

In response, we formulate the problem as a task of feature-topic pairing (\textbf{Figure \ref{fig:motivation}}), which is to align a latent embedding feature space, consisting of multiple latent features, and a textual semantic topic space, consisting of multiple topic labels during SRL.  
The basic idea is to teach a machine to extract topic labels from texts, and then pair the labels with learned features. To that end, we propose to develop a novel deep learning framework to unify feature learning, topic selection, feature-topic matching. 
There are three unique challenges in addressing the problem: (1) \underline{Label Generation Challenge}, in which a textual semantic topic space is difficult to construct due to the unstructured spatial texts;
(2) \underline{Measurement Challenge}, in which a promising measurement is highly desired to evaluate the alignment or quantify the matching score between the topic label space and the embedding feature space; 
(3) \underline{Optimization Challenge}, in which a deep optimization framework is needed for to jointly and simultaneously unify the three tasks of feature learning, topic label selection, and feature-topic pairing.

\begin{figure*}[!t]
	\vspace{-0.25cm}
	\centering\includegraphics[width=0.9\linewidth]{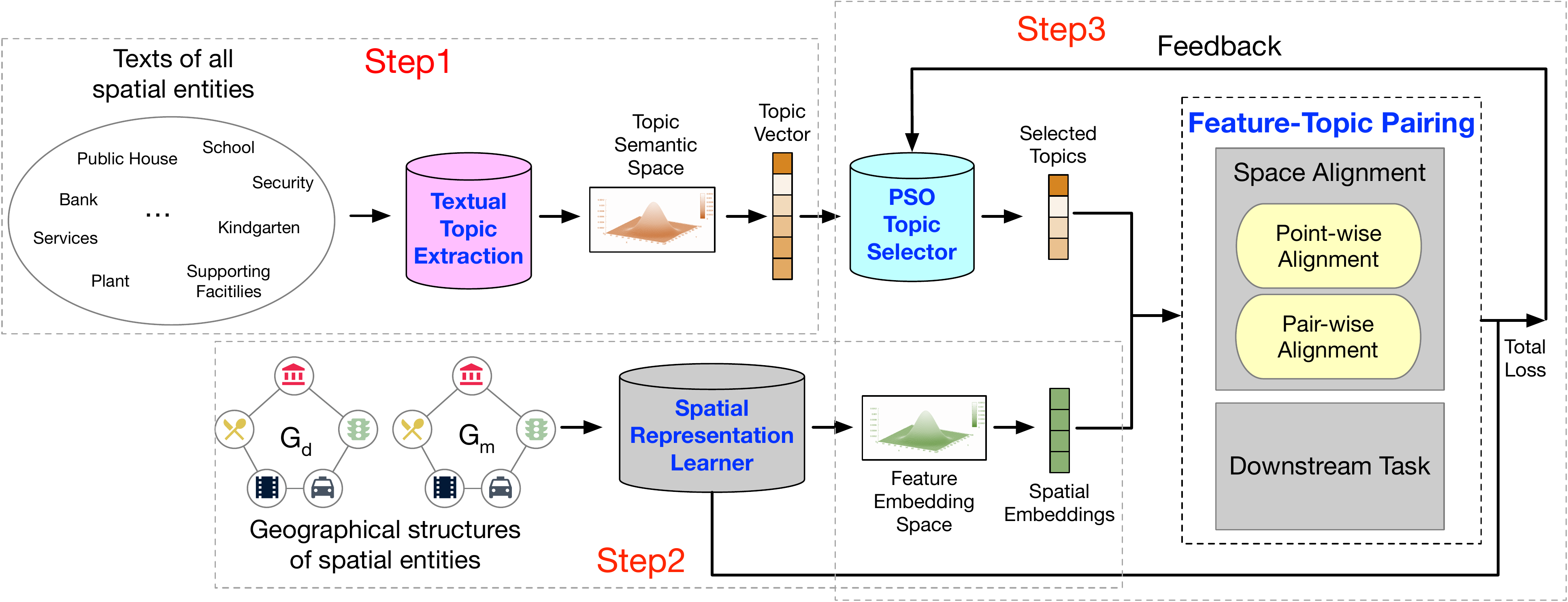}
	\vspace{-0.2cm}
	\caption{An overview of AutoFTP. 
	In the framework, we first construct a topic semantic space based on the texts of spatial entities.
	Then, we initialize a embedding feature space based on the geographical structures of spatial entities.
	Later, we employ a PSO-based framework to conduct feature-topic pairing through jointly optimizing representation learning, point-wise alignment, pair-wise alignment, and downstream task over learning iterations.}
	\label{fig:framwork}
	\vspace{-0.5cm}
\end{figure*}

To solve the three challenges, we develop a new PSO-based framework (named AutoFTP) that enclose the optimizations of feature learning,  topic selection, and feature-topic pairing in a loop. 
Specifically, our contributions are: 
(1) formulating the feature-topic pairing problem for relieving the scarcity of semantic labels; 
(2) proposing a three-step method for generating candidate topic labels; 
(3) deriving a feature-topic alignment measurement by point-wise alignment  between an embedding feature vector and a categorical topic distribution, and pair-wise alignment for the consistency of feature-feature similarity matrix and topic-topic similarity matrix;
(4) developing a Particle Swarm Optimization (PSO)-based algorithm for unified optimization.

\vspace{-0.2cm}
\section{Proposed Method}

\vspace{-0.1cm}
\subsection{The Feature-Topic Pairing Problem}
The feature-topic pairing problem aims to pair the latent features extracted by representation learning, with the explicit topics of texts of a spatial entity. 
Formally, given a set of $N$ spatial entities, the $n$-th entity is described by multiple graphs (e.g., a POI-POI distance graph $G_n^d$ and  a POI mobility connectivity $G_n^m$, defined in Section 3.3) and a topic distribution $\mathbf{t}_n$ extracted from textual descriptions $\mathcal{E}_{n}$. Let $\mathbf{\tilde{r}}_n$ be the embedding vector of the n-th entity.
The objective is to optimize a function that measures representation loss and feature-topic alignment:
\begin{equation}
\textbf{argmax}_{\Tilde{R}} \sum_{n=1}^{N} f(\mathbf{\tilde{r}}_n |\mathbf{t}_n, G_n^d, G_n^m, K\}, 
\end{equation}
where $\mathbf{\tilde{R}} = \{\mathbf{\tilde{r}}_n \}_{n=1}^N \in \mathbb{R}^{N \times K}$ are the embeddings of all spatial entities, $K$ is the number features of an embedding vector.

\vspace{-0.2cm}

\vspace{-0.1cm}
\subsection{Preprocessing}
\noindent \underline{Textual Topic Extraction.}
We employ a pre-trained deep word embedding model~\cite{hanlp} to generate topics. 
Specifically, we first collect the text descriptions of all entities.
Besides, we extract keywords from texts using the TextRank algorithm ~\cite{mihalcea2004textrank} and leverage a pre-trained language model~\cite{hanlp} to learn the corresponding word embedding of each keyword. Moreover, we exploit a Gaussian Mixture Model (GMM) to cluster the keyword embeddings into $T$ topics. 
The clustering model provides a topic label for each keyword.

\noindent \underline{Embedding of Spatial Entities.} 
We construct a graph to capture the spatial autocorrelation between spatial entities. 
Specifically, we describe a spatial entity in terms of its POIs, by building two graphs. (i)
\textbf{POI-POI distance graph:} denoted by $G^d$,  where  POI categories are nodes and the average distances between POI categories are edge weights.
(ii) \textbf{POI-POI mobility graph:} denoted by $G^m$, where nodes are POI categories, and edge weights are human mobility connectivity, which is extracted by the method in~\cite{wang2018learning}. 
We then apply Graph Auto Encoder (GAE)~\cite{kipf2016variational} as the spatial representation learner to learn spatial embeddings over these two constructed graphs respectively. 
Finally, we aggregate the embeddings of these two graphs by avaraging, so as to construct the unified spatial embedding of the entity, denoted by $\mathbf{r}_n \in \mathbb{R}^K$.

\vspace{-0.1cm}
\subsection{PSO Based Feature-Topic Pairing }

\medskip
\noindent\textbf{2.4.1 Measuring the Alignment of Embedding and Semantic Spaces.} 
To pair features with topics, we conduct space alignment from the point-wise and pair-wise perspectives, with considering the alignment of the coordinate system and information contents respectively.
% Referring to definitions 2.2 and 2.3, we aim to align the topic semantic space and feature embedding space from the coordinate system and information contents respectively.
% During the aligning process, we minimize the point-wise alignment loss $\mathcal{L}_P$ and pair-wise alignment loss $\mathcal{L}_C$.
To be convenient, we take the $n$-th entity as an example to explain the calculation process.

% Taking the spatial embedding of the $n$-th entity as an example.
% We first pick a subset of $K$ topics from the topic semantic space.
% Then, we update the spatial embedding vector $\mathbf{r}_n$ for aligning the selected $K$ topics by minimizing the point-wise alignment loss $\mathcal{L}_P$ and pair-wise alignment loss $\mathcal{L}_C$:
% then develop an optimization method to guide the spatial representation leaner to learn an embedding vector $\mathbf{r}_n$ that is aligned with the $K$ topics by 

\underline{\textit{1) Point-wise Alignment Loss: $\mathcal{L}_P$.}} 
Intuitively, the embedding feature of the spatial entity and corresponding topic should reach a consensus on describing an spatial entity, thus correlations are expected to be maximized between them.
Therefore, we first select $K$ values from the topic vector $\mathbf{t}_n$ as the vector $\mathbf{\check{t}}_n \in \mathbb{R}^K$, which contains the most representative semantics in the semantic space.
Then, we maximize the correlation between $\mathbf{\check{t}}_n$ and the spatial embedding $\mathbf{r}_n$, which is equal to minimize the negative correlation between the two vectors.
The formula of the minimizing process as follows:
\begin{equation}
\mathcal{L}_{P} = -\sum_{n=1}^N \frac{\text{cov}(\mathbf{\check{t}}_{n} ,\mathbf{r}_{n})}{\delta (\mathbf{\check{t}}_n) \delta (\mathbf{r}_n)},
\label{lp}
\end{equation}
where $\text{cov(.)}$ denotes the covariance calculation;
$\delta(.)$ denotes the standard deviation.

% Minimizing the formula \eqref{lp} equals to 
% maximize the Pearson Correlation Coefficient between $\mathbf{t}_n$ and $\mathbf{r}_n$.
% The goal is to minimize $\mathcal{L}_P$ during the optimization process.

\underline{\textit{2) Pair-wise Alignment Loss: $\mathcal{L}_C$.}} 
On the other hand, the embedding feature and the corresponding topic should show consistency on the pair-wise similarity in each space to reflect the pair-wise alignment. 
Therefore, we minimize the difference between the pair-wise similarity between these two spaces.
Specifically, we first construct the topic-topic similarity matrix $\mathbf{S}$ and the feature-feature similarity matrix $\mathbf{S}'$.
Specifically, for $\mathbf{S} \in \mathbb{R}^{K \times K} $, we calculate the similarity between any two topics.
For $\mathbf{S}' \in \mathbb{R}^{K \times K}$, we calculate the similarity between two features of spatial embeddings.
We keep the pair-wise consistency between $\mathbf{S}$ and $\mathbf{S}'$ by minimizing the Frobenius norm, as follows:
\begin{equation}
\mathcal{L}_{C} =  ||\mathbf{S} -\mathbf{S}'||_F .
\end{equation}

% \vspace{-0.3cm}

% We hope to minimize $\mathcal{L}_C$, which ensures each dimension in $\mathbf{r}_k$ to represent each topic in $p$ topics.

% \textbf{\color{red} need a figure to illustrate the iterative closed-loop updating process. way too complicated}
% \medskip
\noindent\textbf{2.4.2 Supervised PSO For Automatic Topic Selection.}
To select best K topics for feature-topic alignment, we propose to formulate the joint task of feature learning, topic selection, topic and feature pairing into a PSO problem. 
Specifically, we first randomly initialize a number of particles in PSO, where a particle is a binary topic mask (i.e.,  the mask value of 1 indicates ``select'' and the mask value of 0 indicates ``deselect''). In other words, a set of particles select a subset of topics. 
A multi-objective deep learning model, whose objective function includes the losses of  graph reconstruction, semantic alignment, and the regression estimator in the downstream task, is trained to learn spatial representations, using each selected topic subset.
As an application, we use the embedding of spatial entities (residential communities) to predict their real estate prices, and the loss of the regression model $\mathcal{L}_{Reg}$ is:
\begin{equation}
    \mathcal{L}_{Reg} = \frac{1}{N}\sum_{n=1}^N (c_n-c^*_n)^2,
\end{equation}
where $c_n$ is the golden standard real estate price and $c^*_n$ is the predicted price.
Next, we calculate the fitness of each particle according to the total loss of the deep model. 
The fitness can be calculated by:
\begin{equation}
Fitness = \mathcal{L}_C+\mathcal{L}_P+\mathcal{L}_R+\mathcal{L}_{Reg}. 
\end{equation}
Then, we utilize the fitness to inform all particles how far they are from the best solution.
Next, each particle moves forward to the solution based on not only its current status but also all particles' movement.
After the fitness value of PSO converges, PSO identifies the best topic subset.
Finally, the semantically-rich embeddings of spatial entities, given by: 
$\mathbf{\tilde{R}} = \{\mathbf{\tilde{r}}_n \}_{n=1}^N$.

\vspace{-0.2cm}
\section{Experimental Results}
\subsection{Evaluation Task}
In this work, we apply the proposed AutoFTP to the price prediction of real-estate as the evaluation task. 
Specifically, we first apply AutoFTP to learn a series of representations of spatial entities based on their geographical structural information and related text descriptions.
Then, we build up a deep neural network (DNN) model for predicting average real estate price of each spatial entity according to its corresponding representation. 
We use RMSE, MAE, MAPE and MLSE as the evaluation metric.

\subsection{Data Description}
Table \ref{table:data_stat} shows the statistics of five data sources used in the experiments. 
Specifically, the taxi traces data describes the GPS trajectory of taxis in Beijing in three months; the residential regions, texts, and real estate price data sources are crawled from www.fang.com; and the POIs information are extracted from www.dianping.com.

\begin{table*}[!ht]
\renewcommand\arraystretch{1}
	\centering
	\caption{Overall Performance with respect to RMSE, MAE, MAPE and MSLE. \textbf{(The smaller value is, the better performance is)}}
		\vspace{-0.3cm}
	\begin{tabular}{c|cc|cc|cc|cc}
		\hline
		           & RMSE     & Outperform & MAE     & Outperform & MAPE     & Outperform & MSLE     & Outperform \\ \hline
\textbf{AutoFTP} & \textbf{18.646} & -          & \textbf{16.192} & -          & \textbf{58.851} & -          & \textbf{0.2267} & -          \\
AttentionWalk        & 21.418 & $+14.9\%$  & 19.712 & $+21.7\%$ & 68.590 & $+16.6\%$  & 0.2907 & $+28.2\%$  \\
ProNE        & 21.830 & $+17.1\%$  & 19.929 & $+23.1\%$ & 69.188 & $+17.6\%$  & 0.2949 & $+30.1\%$  \\
GatNE        & 21.229 & $+13.9\%$  & 19.288 & $+19.1\%$ & 67.043 & $+13.9\%$  & 0.2854 & $+25.9\%$  \\
GAE        & 21.338 & $+14.4\%$  & 19.676 & $+21.5\%$ & 68.579 & $+16.5\%$  & 0.2894 & $+27.6\%$  \\
DeepWalk        & 23.561 & $+26.4\%$     & 21.987 & $+35.8\%$  & 76.038 & $+29.2\%$  & 0.3321 & $+46.5\%$   \\
Node2Vec   & 22.688 & $+21.7\%$  & 21.084 & $+30.2\%$  & 73.135 & $+24.3\%$  & 0.3152 & $+39.0\%$  \\
Struc2Vec   & 21.589 & $+15.8\%$  & 19.937 & $+23.1\%$  & 69.423 & $+17.9\%$ & 0.2942 & $+29.7\%$ \\
AutoFTP$^R$   & 21.965 & $+17.8\%$  & 20.283 & $+25.3\%$  & 70.991 & $+20.6\%$ & 0.2928 & $+29.1\%$ \\
AutoFTP$^{(R+P)}$   & 20.509 & $+9.99\%$  & 18.921 & $+16.8\%$  & 66.477 & $+12.9\%$ & 0.2681 & $+18.3\%$ \\
AutoFTP$^{(R+C)}$   & 21.014 & $+12.7\%$  & 19.413 & $+19.8\%$  & 67.920 & $+15.4\%$ & 0.2773 & $+22.3\%$ \\
AutoFTP$^{(R+P+C)}$   & 20.211 & $+8.39\%$  & 18.676 & $+15.3\%$  & 65.685 & $+11.6\%$ & 0.2636 & $+16.3\%$ \\

\hline
	\end{tabular}
	\vspace{-0.2cm}
	\label{tab:overall preformance}
	\vspace{-0.1cm}
\end{table*}

\begin{table}[!thbp]
	\vspace{-0.1cm}
	\centering
	\scriptsize
	\tabcolsep 0.04in
	%	\captionsetup{justification=centering}
	\renewcommand{\arraystretch}{1.3}
	\caption {Statistics of the Experimental Data}
	\vspace{-0.3cm}
	\begin{tabular}[t]{l|l|l}
		\hline
		\textbf{Data Sources}                                   & \textbf{Properties}         & \textbf{Statistics}   \\ \hline
		\multirow{2}*{Taxi Traces}                              & Number of taxis             & 13,597                \\
		& Time period                 & Apr. - Aug. 2012      \\
		\hline
		\multirow{2}*{\tabincell{c}{Residential\\ Regions}}              & \tabincell{c}{Number of residential regions}      & 2,990                 \\
		& Time period of transactions & 04/2011 - 09/2012     \\ \hline
		
		\multirow{2}*{POIs}    & Number of POIs          & 328668    \\
		& Number of POI categories          & 20             \\ \hline
		
		\multirow{2}*{Texts}    & Number of textual descriptions          & 2,990    \\ 
		& Time Period                      & 04/2011 - 09/2012 \\ \hline
		
		\multirow{2}*{Real Estate Prices}    & Number of real estate prices         & 41,753    \\ 
		& Time Period                      & 12/2011 - 06/2012 \\ \hline
	\end{tabular}
	\vspace{-0.55cm}
	\label{table:data_stat}
\end{table}

\vspace{-0.1cm}
\subsubsection{\textbf{ Baseline Algorithms.}}
We compared our proposed method with seven baseline algorithms: \textbf{AttentionWalk} ~\cite{abu2018watch}, \textbf{ProNE} ~\cite{zhang2019prone}, \textbf{GatNE}~\cite{cen2019representation}, \textbf{GAE}~\cite{kipf2016variational}, \textbf{DeepWalk}~\cite{perozzi2014deepwalk}, \textbf{Node2Vec}~\cite{grover2016node2vec}, and

\noindent \textbf{Struc2Vec}~\cite{ribeiro2017struc2vec}. Besides, regarding the are four losses in AutoFTP: reconstruction loss $\mathcal{L}_R$,
point-wise alignment loss  $\mathcal{L}_P$,
pair-wise alignment loss $\mathcal{L}_C$, and regression loss $\mathcal{L}_{Reg}$., we also derive four variants: (ii) $\textbf{AutoFTP}^{(R+P)}$, which keeps $\mathcal{L}_R$ and $\mathcal{L}_P$ of AutoFTP;
(iii) $\textbf{AutoFTP}^{(R+C)}$, which keeps $\mathcal{L}_R$ and $\mathcal{L}_C$ of AutoFTP;
(iv) $\textbf{AutoFTP}^{(R+P+C)}$, which keeps $\mathcal{L}_R$, $\mathcal{L}_P$, and $\mathcal{L}_C$ of AutoFTP.

\vspace{-0.35cm}
\subsection{Overall Performance}
Table \ref{tab:overall preformance} shows the comparison of all the 11 models. As can be seen, AutoFTP, in overall, outperforms the baseline algorithms in terms of RMSE, MAE, MAPE and MSLE.
A possible reason for this observation is that compared with other baseline algorithms, AutoFTP not just captures geographical structural information but also preserves rich semantics of spatial entity.
Besides, the regression estimator (the downstream task) of AutoFTP provides a clear learning direction (accuracy) for spatial representation learning.
Thus, in the downstream predictive task, the spatial embedding features learned by AutoFTP beats all baselines.

\section{Related Work}
% \textbf{Urban Computing.}
% Our work is closely related to urban computing.
% Urban computing~\cite{zheng2014urban} refers to the study of analyzing and modeling urban data (e.g., traffic flow, human mobility, and geographical data).  Yuan et al. discover regional functions of a city using POIs and taxi traces \cite{yuan2012discovering}.
% Work \cite{ceci2007discovering} identifies emerging patterns with multi-relational approach from spatial data. 

\textbf{Graph Representation Learning with Latent Semantics.} Graph representation learning refers to techniques that preserve the structural information of a graph into a low-dimensional vector ~\cite{wang2020exploiting,wang2016structural}. 
% For instance, Kipf {\it et al.} obtained graph embedding through an encoder-decoder paradigm by minimizing the reconstruction loss ~\cite{kipf2016variational}.
% Peng {\it et al.} implemented graph representation learning via maximizing the graphical mutual information ~\cite{peng2020graph}.
% Zhang {\it et al.} enhance the graph embeddings by propagating them in the spectrally modulated space ~\cite{zhang2019prone}. 
However, owing to traditional graph representation learning models are implemented by deep neural networks, the learned embeddings lack interpretability.
Recently, to overcome this limitation, researchers leveraged the texts related to graphs to learn semantically rich representations~\cite{mai2018combining,xiao2017ssp}. 

\noindent\textbf{Topic Models in Spatio-temporal Domain.}
Topic models aim to automatically cluster words and expressions patterns for characterizing documents ~\cite{xun2017correlated,lee2018identifying}.
% For instance, LDA~\cite{blei2003latent} is a classical topic model, which utilizes bag-of-words to calculate topic distribution.
% Ida2vec~\cite{moody2016mixing} utilizes word embedding and topic model to produces interpretable document representation.
Recently, to understand the hidden semantics of spatial entities, many researchers applied topic models in the spatio-temporal data mining domain ~\cite{huang2020mobility,huang2019adaptive}. 
% For instance, Zhao {\it et al.}
%  discovered representative and interpretable human activity patterns from transit data automatically by a spatio-temporal topic model \cite{zhao2020discovering}.
% Yao {\it et al.} tracked spatio-temporal and semantic dynamics of urban geo-topics based on an improved dynamic topic model that embeds spatial factors of pairwise distances between tweets \cite{yao2020tracking} .
% These successful applications validate the effectiveness of topic models for extracting semantics in spatio-temporal domains.
% However, traditional topic models only focus on word frequency in texts but neglect the semantics of words.
% Recently, the success of many pre-trained language models \cite{kenton2019bert,vaswani2017attention,yang2019xlnet} brings hope for producing more reasonable topic distribution.
Thus, in this paper, we employ a pre-trained language model to get the embeddings of keywords and utilize Gaussian Mixture Model to extract topic distribution based on the embeddings.

\vspace{-0.2cm}
\section{Conclusion}
We presented a novel spatial representation learning (SRL) framework, namely AutoFTP.
The spatial embeddings produced by traditional SRL models lack semantic meaning.
To overcome this limitation, we formulated the feature-topic paring problem. 
We proposed a novel deep learning framework to unify representation learning, topic label selection, and feature-topic pairing through a PSO-based optimization algorithm.
% Specifically, 
% %we first leverage the texts of spatial entities to extract a candidate topic label set, and, thus, construct a topic semantic space.
% we designed a segmentation-embedding-clustering method to generate candidate feature topic labels from texts.
% %Then  we initialize a feature embedding space based on the geographical spatial relation of spatial entities.
% We developed an integrated measurement to measure the pointwise and pairwise alignment between topic label and embedding feature space. 
% %After that, we employ a PSO-based framework to optimize the topic selection and feature-topic pairing automatically.
% We devised a PSO based optimization algorithm to effectively solve the joint task of feature learning and feature-topic pairing.
% %During the optimization, the topic semantic space and the feature embedding space are aligned from point-wise and pair-wise perspectives.
% Our method integrated spatial graphs and associated texts to learn effective embedding features with visible labels.
% %Eventually, the learned spatial feature embeddings not only capture the geographical information but also grab the semantic meaning.
Extensive experiments demonstrated the effectiveness of AutoFTP by comparing it with other baseline models. 
% The topic labels of the learned features were shown by many case studies and the feature importance analysis of a downstream task.
For future work, we plan to extend our approach from geospatial networks to other applications that consist of graphs and texts, such as social media and software code safety.

\section*{Acknowledgments}
This research was partially supported by the National Science Foundation (NSF) via the grant numbers: 1755946, 2040950, 2006889, 2045567, 2141095.

\bibliographystyle{ACM-Reference-Format}
\bibliography{acm}

\end{document}